\documentclass[letter,12pt]{article}

\usepackage{todonotes}

\usepackage{newtxtext,newtxmath, amsfonts, bm,mathtools}

\mathtoolsset{showonlyrefs=true}

\usepackage[utf8]{inputenc}

\usepackage[boxed,lined,linesnumbered]{algorithm2e}

\usepackage{amsmath, textcomp, paralist, bm, natbib}

\usepackage{array}

\usepackage{wrapfig}

\usepackage{multirow}

\usepackage{tabularx}

\usepackage{booktabs}

\usepackage{float, subfig}

\DeclareMathAlphabet{\mathcal}{OMS}{cmsy}{m}{n}

\graphicspath{{Graphics/}}

\usepackage{arydshln}

\usepackage{xcolor}

\usepackage[hyphens,spaces,obeyspaces]{url}

\usepackage{indentfirst}

\usepackage{pdflscape}

\usepackage[flushleft]{threeparttable}

\usepackage{pifont}

\usepackage[margin=3cm]{geometry}

\usepackage[margin=2cm]{caption}

\usepackage{titling}

\usepackage{tikz}

\usetikzlibrary{positioning}

\SetKwInOut{Parameter}{parameter}

\usepackage{multirow}

\usepackage{adjustbox}

\newcolumntype{R}[2]{%
    >{\adjustbox{angle=#1,lap=\width-(#2)}\bgroup}%
    l%
    <{\egroup}%
}
\newcommand*\rot{\multicolumn{1}{R{45}{1em}}}

\usepackage{array, ragged2e}

\newcolumntype{P}[1]{>{\raggedright\arraybackslash}p{#1}}

\begin{document}

\title{DEFT: A new distance-based feature set for keystroke dynamics}

\author{Nuwan Kaluarachchi\textsuperscript{1,2}, Sevvandi Kandanaarachchi\textsuperscript{2}, Kristen Moore\textsuperscript{2}  Arathi Arakala \textsuperscript{1} }

\date{ \scriptsize{October 6, 2023}}

\maketitle
\textsuperscript{1} Mathematical Sciences, STEM College, RMIT University, Melbourne VIC 3000, Australia\\ 
\textsuperscript{2} CSIRO's Data61, Research Way, Clayton VIC 3168, Australia.



* Corresponding author: nuwan.kaluarachchi@student.rmit.edu.au

\begin{abstract}
Keystroke dynamics is a behavioural biometric utilised for user identification and authentication. We propose a new set of features based on the distance between keys on the keyboard, a concept that has not been considered before in keystroke dynamics. We combine flight times, a popular metric, with the distance between keys on the keyboard and call them as Distance Enhanced Flight Time features (DEFT). This novel approach provides comprehensive insights into a person's typing behaviour, surpassing typing velocity alone. We build a DEFT model by combining DEFT features with other previously used keystroke dynamic features. The DEFT model is designed to be device-agnostic, allowing us to evaluate its effectiveness across three commonly used devices: desktop, mobile, and tablet. The DEFT model outperforms the existing state-of-the-art methods when we evaluate its effectiveness across two datasets. We obtain accuracy rates exceeding 99\%\ and equal error rates below 10\%\ on all three devices.
\end{abstract}


\section{Introduction}
Keystroke dynamics refers to the systematic analysis of the pattern of key press and release on a physical or virtual keyboard by an individual. The information distilled from the typing patterns enables user identification and authentication. Thus, it has proven to be an effective behavioural biometric modality. One notable advantage of keystroke dynamics is its suitability for continuous authentication, as it involves an ongoing behaviour performed by an individual throughout a typing session. This continuous nature makes it well-suited for establishing and maintaining user authentication in various contexts.


 Most keystroke dynamics studies in the last five years used temporal features \textbf{(TEMP)} \cite{kasprowski2022biometric,acien2021typenet,yang2021tkca,yang2021fktan,aversano2021continuous,kiyani2020continuous, kim2020freely, ayotte2019fast}. TEMP features compute uni-graph (key hold time), digraph (flight times) and trigraph (presses of three consecutive keys) attributes from typing behaviours. Alsultan et al. \cite{alsultan2017non} used non-conventional features \textbf{(NC)} such as average backspace and negative flight times over a user typing session. Al-Saraireh and AlJa'afreh \cite{al2023keystroke} and Belman and Phoha \cite{belman2020discriminative} used flight times of commonly typed keypairs \textbf{(CKP)} as features. While either TEMP, NC, or CKP features have been used individually in previous studies, their combined effect has not been explored.

Our study introduces a new set of features that considers the distance between key pairs when considering flight times. We call these Distance Enhanced Flight Time \textbf{(DEFT)} features. We combine DEFT with the previously studied TEMP, NC and CKP features (DEFT Model) and investigate the performance of the combined feature for continuous authentication. We show that using DEFT features significantly improves authentication performance on desktop, mobile and tablet compared to existing approaches. In addition, most studies only focus on free text. However, we frequently type usernames, passwords and email addresses, which come under the fixed text category. So in this study, we present our results with fixed and free text-typing datasets to incorporate the mixed nature of typing. Furthermore, while most device authentication studies are limited to a single device, with a handful of studies using two devices \cite{belman2020doubletype}, we demonstrate the broad applicability of this new set of features on three commonly used devices: desktop, mobile and tablet.

\section{Related Work}
Over the last 5 years, various feature types and classifiers have been tested for user identification and authentication with keystroke dynamics. While a detailed review of this work is out of scope, we briefly review (see Table \ref{tab:litsurvey}) some insights gained from this study and explain how it has impacted our research. As seen from Table \ref{tab:litsurvey}, most studies compute features using flight times, hold times, words per minute and error rate. In contrast, Alsultan et al. \cite{alsultan2017non} consider a novel set of features that capture the backspace, negative flight times and shift key usage in a session.  A key takeaway from Table \ref{tab:litsurvey} is that none of the recent studies has combined these different types of features to construct an optimal set of features for user identification and authentication.

\begin{table}[!ht]
\begin{tabular}{r|cccccccc}
\hline
\hline
Year [Ref]
&
\rot{Hold Times} &
\rot{Flight Times} &
\rot{WPM} &
\rot{Error Rate} &
\rot{NegUD} &
\rot{NegUU} &
\rot{Shift Usage} &
\rot{Cpslck Usage} \\
\hline
2022 \cite{kasprowski2022biometric}     &X  &X  &  &  &  &  &   &      \\ 
2022 \cite{acien2021typenet}            &X  &X  &  &  &  &  &   &    \\
2021 \cite{yang2021tkca}                 &X  &X  &  &  &  &  &   &    \\       
2021 \cite{yang2021fktan}                &X  &X  &  &  &  &  &   &    \\
2021 \cite{aversano2021continuous}        &  &X  &X  &X  &  &  &   &    \\
2020 \cite{kiyani2020continuous}         &X  &X  &  &  &  &  &   &    \\
2020 \cite{lu2020continuous}             &X  &X  &  &  &  &  &   &    \\
2020 \cite{belman2020discriminative}     &X  &X  &  &  &  &  &   &    \\
2020 \cite{kim2020freely}                 &X  &X  &  &  &  &  &   &    \\
2019 \cite{ayotte2019fast}               &X  &X  &  &  &  &  &   &    \\
2019 \cite{wu2019user}                    &X  &X  &  &  &  &  &   &    \\
2017 \cite{alsultan2017non}               &  &  &X   &X  &X  &X  &X   &X    \\
2017 \cite{mondal2017person}               &X  &X  &  &  &  &  &   &    \\
\hline
\end{tabular}
\caption{Comparison of feature types used in recent keystroke dynamic studies. The majority of studies used Temporal features, while only one study used only non-conventional features for keystroke dynamics. None of the studies used a combination of all these feature types.}
\label{tab:litsurvey}
\end{table}

We combine these different types of features, and additionally, we develop a new set of features based on the physical distance between keyboard keys, which we call Distance Enhanced Flight Time (DEFT) features. The intuition behind these features is that flight and hold times depend on the distance between keys and the use of one or both hands. For example, if we consider keys typed by a single hand, we expect a normal user to take more time between two keys that are further apart, such as 'A' and 'T', compared to the time taken for two keys that are closer such as 'A' and 'S'. Subsequently, we employ feature selection strategies to determine whether the DEFT features contribute to discriminating between users. 


\section{Methodology}
\subsection{Datasets}

We use the publicly available SU-AIS BB-MAS (Syracuse University and Assured Information Security Behavioral Biometrics Multi-device and multi-Activity data from Same users) dataset \cite{datset} as our main dataset. This dataset plays a vital role in advancing the field of biometrics, addressing the existing gap in collecting data from individuals across multiple devices. The BB-MAS dataset captures multiple behavioural biometric modalities from three different devices, including swiping, keystroke, and gait dynamics. The data was collected over two months from 117 participants. However, missing data in the collection process leaves us with only 116 users for each device. A comprehensive account of the dataset and the data collection process can be found in \cite{belman2019insights}. 

We evaluate our newly introduced DEFT features using another well-known keystroke dynamic dataset, the Buffalo dataset \cite{sun2016shared}, to validate its efficacy. This dataset consists of keystroke data collected from 148 users using 4 different types of keyboards. The keystroke patterns from 75 users were captured by their typing on the same type of keyboard in 3 distinct sessions. Keystroke patterns from the other 73 users were acquired by typing on three keyboards in the 3 different sessions.

\subsection{Distance Enhanced Flight Times (DEFT) features}
\begin{figure}[htb]
\centering
    \includegraphics[scale=0.45]{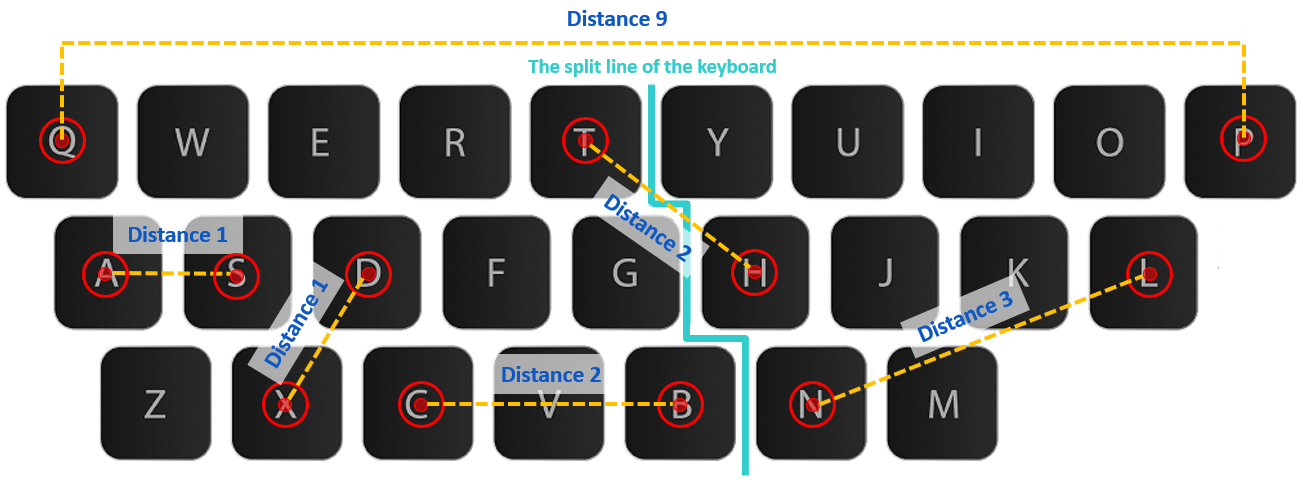}
    \caption{\label{keyboardLayout} Calculation of the distance between the key pairs on the keyboard. $<$A,S$>$ is a distance 1 digraph while $<$T,H$>$ is a distance 2 digraph and $<$N,L$>$ is a distance 3 digraph. The longest distance between a key pair is distance 9. The blue line separates the keyboard into left and right sides, which helps identify keys typed by an individual's left or right hand.
}
\end{figure}

DEFT features are computed under the assumption that the distance between the keys affects the flight time. The distance is computed using the spatial separation of the keys on the keyboard (see Figure \ref{keyboardLayout}). For example,  we expect the flight times between keys, such as `A,S', `S,D' and `W,E', to be similar (generated from a single probability distribution) due to the equal distance between the respective keys. The distance between the keys in each of those 3 pairs is 1. We consider the average flight time for all such key pairs, i.e., the average flight time (per user) across all distance 1 key pairs. Similarly, we calculate the average flight time for distances 0, 2, and 3 key pairs. In this manner, the DEFT features augment the flight times (TEMP features) by incorporating the distance between keys.




Figure \ref{keyboardLayout} demonstrates the distances between some key pairs.  The calculated key pair distances range from zero to nine, with zero indicating the pressing of the same key and nine denoting the longest distance between two keys, specifically,  `Q' and `P'. Generally, people use both hands for typing; they use the left hand to type keys on the left and the right hand for keys on the right.  We separate the left and right sides of the keyboard as demarcated by the blue line in Figure \ref{keyboardLayout}. If both keys of the digraph are on the left side, we indicate it with LL. If both keys are on the right side, it is indicated with RR; if both digraph keys span either side of the keyboard, we denote it by LR.

Next, we combine the distance between the keys with the flight times. In the study, we try to identify the typing patterns of a single user for each hand. Therefore, using the standard QWERTY finger placement, we consider the maximum distance between keys typed by a single hand to be three units. Consequently, flight times for distances ranging from zero to three on both the left and right sides of the keyboard are selected for analysis. We focus on LL and RR key pairs and disregard LR key pairs. We compute the average of each of the four flight times, F1, F2, F3 and F4, as declared by Belman and Phoha in \cite{belman2020discriminative, belman2020doubletype} grouping by the distance between the keys for each hand. Thus, we have 32 ( $4\times 4 \times 2$) new features ranging from F1\_distance\_0\_LL to F4\_distance\_3\_RR. These are the DEFT features, and they capture the average distance flight times per distance for a given user.  

Unlike existing text-based features in the keystroke dynamics, which mainly revolve around temporal variations, including instances of key press and release, temporal gaps, and the frequency of such events, an aspect formerly unexplored involves the measurement of spatial separation between pairs of keys. Addressing this gap, our study introduces a novel approach by incorporating key pair distances in conjunction with the flight times of said key pairs. This innovative amalgamation enhances the depth and breadth of our analysis, ushering in a more comprehensive understanding of keystroke dynamics.


\begin{figure}[!htb]

\centering
    \includegraphics[ scale = 0.35]{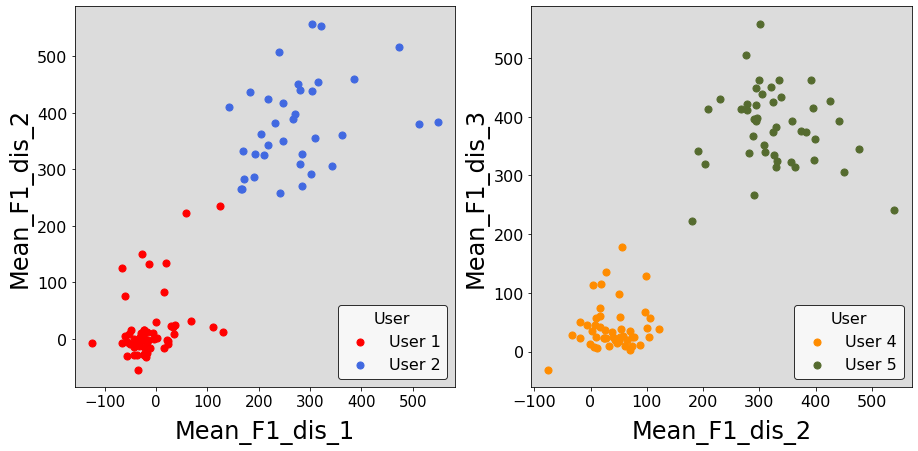}
    \caption{\label{CompareDistances} The pairwise DEFT features of four different users for key pairs typed by the left hand on desktop devices. The average flight 1 (F1) timing for the two users for distance 1 key pairs and distance 2 key pairs is shown in the left figure, and the average flight 1 (F1) timing for two different users for distance 2 key pairs and distance 3 key pairs is shown by the right figure. The figures imply the validity of separating the flight times based on the key pair distances for user identification. Each point of a certain colour represents a user sample.}
\end{figure}

Figure \ref{CompareDistances} shows the pairwise distribution of three DEFT features for four different users.  The results for these two examples illustrate distinct user clusters, which illustrates that the features  (F1\_distance\_1\_LL, F1\_distance\_2\_LL and F1\_distance\_3\_LL) are user dependent and can help user identification and authentication. As the distances between keys and flight times can vary depending on the type and size of the device, such as desktop, mobile or tablet, we adopt a solution that yields relative values for the distances. This is achieved by normalising the distances through division by a single key size, ensuring that the derived distances remain independent of the device being used. This device-agnostic approach can accurately capture and compare the relative distances between keys, irrespective of device type or screen size variations.

\section{Results and Discussion}\label{sec:results}
\subsection{Feature selection}\label{sec:featureselection}
We compute TEMP, NC, CKP and DEFT features for each user sample in the BBMAS dataset, resulting in an expanded set of keystroke dynamic features. We found that the most frequently typed key pairs in the dataset were `T,H', `I,S', `H,E', `A,P', `L,E' and `C,O'. When calculating the flight times for TEMP, CKP and DEFT features, we use a simple filter to detect and remove the high or low time differences in keystroke dynamics. We eradicate any instances of a time difference of more than five seconds. We assume these scenarios happen by pauses, getting instructions during the data collection or recording issues. After calculating the TEMP, NC, CKP and DEFT features, we get the average values of each feature for each sample. 
To identify the most discriminative features from the expanded feature set, we employ the Random Forest (RF) classifier, specifically tuned for multi-class classification described by Ayotte \cite{ayotte2019fast}. To conduct our analysis, we split the dataset into a training set comprising 70\%\ of the user samples and a testing set comprising the remaining 30\%\. This process is carried out separately for desktop, mobile, and tablet devices. This procedure identifies 37, 41, and 42 discriminative features for each device, respectively. Table \ref{tab_SF} shows the types and number of the shortlisted features for all three devices. Notably, the table reveals that the DEFT features exhibit the highest occurrence rate among the shortlisted features, accounting for more than 50\%\ of the selected features in the case of mobile and tablet devices. This finding demonstrates the effectiveness of the DEFT features in capturing discriminative information necessary for keystroke authentication.

\begin{table}[htb]
\centering
\begin{tabular}{llll}
Feature Category & Desktop & Mobile & Tablet\\
\hline\hline
\textbf{DEFT} & \textbf{17} & \textbf{25} & \textbf{23}\\
 CKP & 9 & 8 & 10 \\
TEMP & 6 & 6 & 6 \\
NC & 5 & 2& 3\\
\hline
Total & 37 & 41 & 42\\
\hline
\end{tabular}
\caption{\label{tab_SF} The category breakdown of features with the highest discriminative characteristics selected by the Random Forest classifier. More than 45\%\ of the selected features on desktop and 50\%\ of the selected features on mobile and tablet are DEFT features, showing the dominance of DEFT features in the feature list.}
\end{table}



\subsection{Authentication Framework}
After selecting the discriminative features, we build a binary classifier for each user. A user has about 50 samples from their keystroke data and  6000 samples from other users' keystroke data. Each user's sample is 100 keystrokes long and comprises a vector of the features shortlisted by the feature selection stage. We use stratified five-fold cross-validation with the Extreme Gradient Boost (XGB) classifier. Due to the extreme class imbalance of the mated compared to non-mated samples, we use the Synthetic Minority Oversampling Technique (SMOTE) \cite{chawla2002smote} to oversample the genuine user's class in each fold. 


\begin{figure}[htb]
\centering
    \includegraphics[width=\linewidth ]{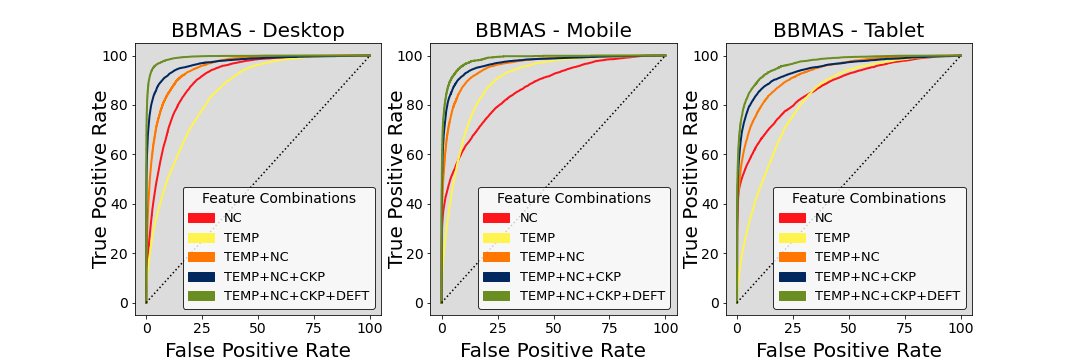}
    \caption{\label{Compareallfeatures} The ROC curve representation of the authentication performance of different combinations of keystroke features for different devices. In all three devices, the keystroke dynamic performance of the models increased by adding the DEFT features.}
\end{figure}

First, we perform an ablation study to validate the discriminative power of our new DEFT features. Figure \ref{Compareallfeatures} shows ROC (Receiver Operating Characteristics) curves for five feature combinations: NC, TEMP, TEMP + NC, TEMP + NC + CKP, TEMP + NC + CKP + DEFT. The feature selection process, as described in Section \ref{sec:featureselection}, is only used for the TEMP + NC + CKP  and TEMP + NC + CKP + DEFT combinations. The final curated set of features after adding DEFT achieves the best performance for all devices. We work with this set of features for the remainder of the paper, which we name the DEFT model.

\begin{figure}[htb]
\centering
    \includegraphics[scale =0.35 ]{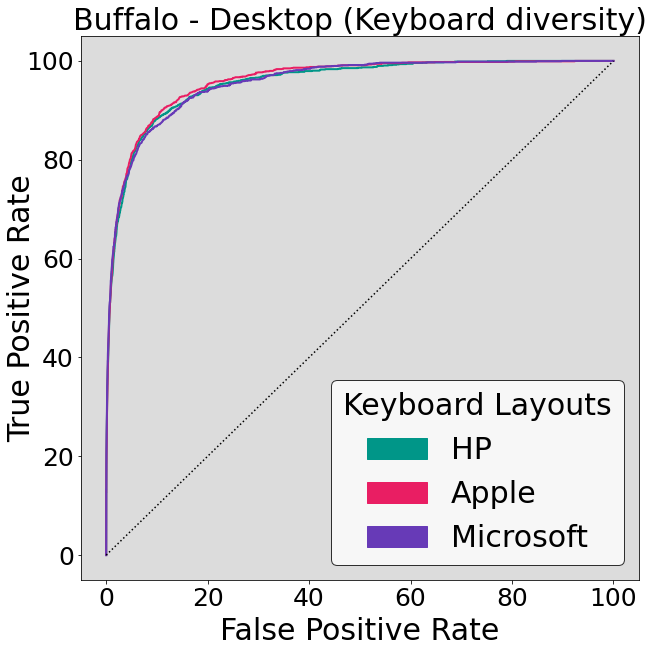}
    \caption{\label{ComparedifferentKeyboards} The ROC curve representation of the authentication performance of the DEFT model on different desktop keyboard types in the Buffalo dataset. The proximity of the curves demonstrates equivalent performance for the three keyboard types.}
\end{figure}
Next, we test if there is a variation in the performance of the DEFT model when the keyboard type changes. In particular, we want to test if disparities in performance arise in the distance-based features when confronted with different keyboard types. The ROC curves on testing the three keyboard types: An HP wireless keyboard, a Microsoft ergonomic keyboard and an Apple wireless keyboard. The results are shown in Figure \ref{ComparedifferentKeyboards}. This demonstrates that the DEFT model remains notably consistent across different keyboard configurations.

\begin{figure}[htb]
    \includegraphics[width=\textwidth]{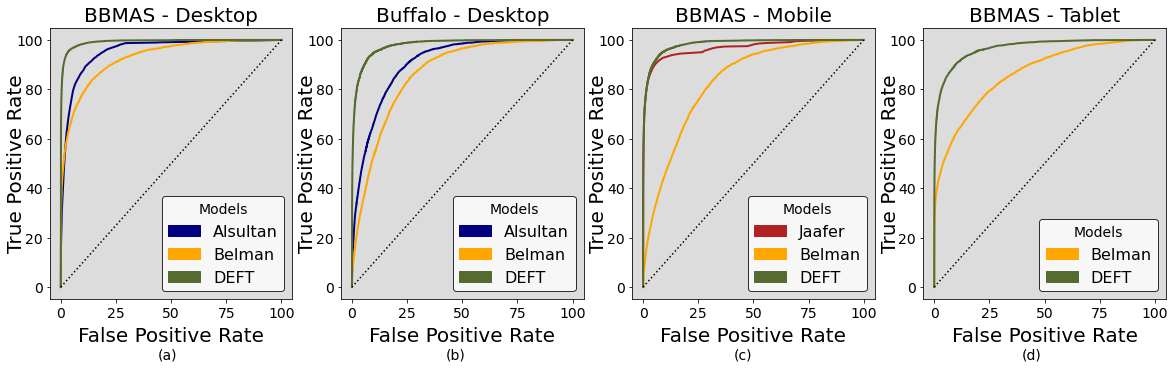}
    \caption{\label{Compareallmodels} Comparison of the authentication performance of the DEFT model against existing keystroke dynamics models across 3 devices and 2 datasets. (a)  The comparison of desktop models on BBMAS. (b) The comparison of desktop models on the Buffalo dataset. (c) The comparison of mobile models on BBMAS. (d)The comparison of tablet models on BBMAS. The DEFT model outperforms other models for all three devices.}
\end{figure}

We next compare our model with leading studies on desktop \cite{alsultan2017non, belman2020discriminative},  mobile \cite{al2023keystroke, belman2020discriminative} and tablet \cite{belman2020discriminative}. Alsultan et al. \cite{alsultan2017non} collected NC features using their own dataset of 30 users. Al-Saraireh and AlJa'afreh \cite{al2023keystroke} analysed CKP features from 54 mobile data users of BBMAS. Belman and Phoha \cite{belman2020discriminative} examined user identification across desktops, mobiles, and tablets using CKP features and context-based multi-class classifiers on only 20 users of BBMAS. In order to compare \cite{belman2020discriminative} with our approach, we convert their multi-class classification to a binary classification problem by selecting the highest-performing context-based classifiers. Although these results are published, the experimental settings in each study are different, including training and testing splits, sample sizes, and evaluation metrics. In our comparison, we use the same experimental setting for all methods. Specifically, we consider all 116 users available in the BBMAS dataset for three devices and all 75 users of the Buffalo dataset for desktops. We plot ROC curves for each model by getting the average of each fold in the cross-validation for all users for each device to get a complete picture of the studies under different thresholds. Figure \ref{Compareallmodels} shows the results, demonstrating that the DEFT model performs better than other models for all three devices of BBMAS and desktop devices of the Buffalo dataset.

\begin{table}[htb]
\centering
\begin{tabular}{p{1.5cm}p{1.5cm}p{1.5cm}p{1.5cm}p{1.5cm}p{1.5cm}}
\hline
Device & Model & Accuracy & EER & F1 & AUC-ROC \\
\hline \hline
 & Alsultan & 95.5 (0.25) & 10.9 (0.63) & 52.1 (1.48) & 95.6 (0.40) \\
 Desktop & Belman &  97.9 (0.03) & 15.1 (0.49) & 60.9 (0.98) & 92.9 (0.29) \\
 & DEFT & \textbf{99.6 (0.02)} & \textbf{3.8 (0.42)} & \textbf{77.5 (1.1)} & \textbf{99.3 (0.07)} \\
 \hline
 & Jaafer & 99.4 (0.02)  & 6.7 (0.64)  & 58.6 (1.2)   & 98.2 (0.24) \\
Mobile & Belman & 88.7 (0.24) & 29.8 (0.77) & 46.9 (0.29)  & 83.1 (0.87) \\
 & DEFT & \textbf{99.4 (0.01)} & \textbf{6.6 (0.51)} & \textbf{65.6 (1.2)}  & \textbf{98.4 (0.23)} \\
\hline
Tablet & Belman &  96.0 (0.04) & 40.4 (0.59) & 47.5 (0.9) & 86.4 (0.80) \\
 & DEFT &  \textbf{99.3 (0.02)} & \textbf{9.8 (0.65)} & \textbf{61.8 (1.1)} & \textbf{96.7 (0.42)} \\
\hline
\end{tabular}
\caption{\label{tabfinal} Summary of the performance of the state-of-the-art keystroke models for the three devices using the BBMAS dataset. All the models followed the same experimental setting with their own features and classifiers. Our DEFT model performed better in all three devices for all performance metrics. The table proves the validation of using DEFT features for keystroke dynamics. All values are in percentages, and parenthesis values are the standard deviation}
\end{table}

Table \ref{tabfinal} summarises the results of our model and other compared state of art models using the BBMAS on keystroke dynamics for the three devices under different performance metrics. As implied by the table, our DEFT model has the highest accuracy, F1 score and AUC-ROC. The model's EER (Equal Error Rate) is 3.8\%\, 6.6\%\, and 9.8\%\ for the desktop, mobile and tablets.

To ensure the reproducibility of our research, we have made relevant code snippets \footnote{\url{https://github.com/NuwanYasanga/DEFT}} available online.

One limitation that arose in our analysis is a significant class imbalance issue within the test datasets, where the imposter class significantly outnumbered the genuine user class by a factor of over 100 in most cases. We don't employ any oversampling or undersampling techniques for the testing set, as we oversample the training dataset. It is important to note that inherent biases towards the imposter class primarily drove low F1 scores. Due to the scarcity of genuine user samples within the test set, even a slight deviation from the expected behaviour of genuine users may lead to misclassification as an imposter sample. This limitation is the ground truth in biometric models, with evaluation encompassing the entirety of 116 users within the BBMAS dataset and the complete cohort of 75 users in the Buffalo dataset. Importantly, this analysis is executed without the application of any data filtration or modification to the test dataset, preserving its original structure and characteristics intact.

The results presented in Table \ref{tabfinal}  and Figures \ref{Compareallfeatures},
\ref{Compareallmodels} indicate that the combined pool of features (TEMP + NC+ CKP + DEFT) in the DEFT model
is the most discriminative, resulting in significant performance improvements in all three devices. These results highlight that the DEFT features played a key role in capturing the keystroke dynamics for all three devices.

\section{Conclusion}
This paper introduces DEFT, a new set of distance-based features for keystroke dynamics. By combining DEFT with existing features, we constructed a pool of features called a DEFT model that achieved improved authentication performance for desktop, tablet and mobile devices. Noting that DEFT features accounted for approximately 50\%\ of the discriminative feature set identified by a feature selection process, we demonstrated the utility and broad applicability of DEFT features across devices and across datasets. Our comprehensive analysis identified that spatial features play a significant role in user discrimination and can be effectively employed for continuous user authentication. In future research, we aim to extend the application of DEFT features in cross-device authentication by applying deep transfer learning techniques.




\bibliography{References.bib}

\bibliographystyle{agsm}

\end{document}